# UCell: rethinking generalizability and scaling of biomedical vision models

**Nicholas Kuang[a], Vanessa Scalon[b] and Ji Yu[c,*]**
[a] Vanderbilt University, School of Medicine, Nashville, TN
[b] UConn Health, Center for Regenerative Medicine and Skeletal Development, Farmington CT
[c] UConn Health, Center for Cell Analysis and Modeling, Farmington CT
*Correspondence should be addressed to JY. (jyu@uchc.edu)

The modern deep learning field is a scale-centric one. Larger models have been shown to consistently perform better than smaller models of similar architecture. In many sub-domains of biomedical research, however, the model scaling is bottlenecked by the amount of available training data, and the high cost associated with generating and validating additional high quality data. Despite the practical hurdle, the majority of the ongoing research still focuses on building bigger "foundation" models, whereas the alternative of improving the ability of small models has been under-explored. Here we experiment with building models with 10-30M parameters, tiny by modern standards, to perform the single-cell segmentation task. An important design choice is the incorporation of a recursive structure into the model's forward computation graph, leading to a more parameter-efficient architecture. We found that for the single-cell segmentation, on multiple benchmarks, our small model, UCell, matches the performance of models 10-20 times its size, and with a similar generalizability to unseen out-of-domain data. More importantly, we found that ucell can be trained from scratch using only a set of microscopy imaging data, without relying on massive pretraining on natural images, and therefore decouples the model building from any external commercial interests. Finally, we examined and confirmed the adaptability of ucell by performing a wide range of one-shot and few-shot fine tuning experiments on a diverse set of small datasets. Implementation is available at https://github.com/jiyuuchc/ucell.

## Introduction

Model scaling is one of the most reliable and predictable strategies to improve a deep learning model's accuracy, generalizability, and many other positive benchmarks. Empirical evidence suggested a power law relationship between generalization error and model/dataset size [1]. Scaling of large language models [2] is a particularly striking success story, leading to some to hypothesize that scaling would eventually lead to emergent properties [3] in deep learning models. It is, therefore, of no surprise that biomedical researchers have also put significant efforts into scaling when modeling scientific data. Model size has grown quickly in recent years, with the current state-of-the-art models typically approaching or already surpassing the one-billion-parameter mark [4,5].

Scaling biomedical models, however, has many practical challenges, the least of which is the shortage of training data that is needed for training larger models. Research data are expensive to produce. Pruning and validating research data requires highly specialized domain knowledge. Both would suggest producing industrial-scale datasets at scale being impractical. The hurdle has not stopped the research efforts of course: For example, in microscopy image modeling, which is the focus of this study, the current dominant approach is to pretrain on a massive corpus of natural image dataset [6,7] compiled by the commercial entities from industry, followed by finetuning on microscopy images to introduce specialization [8-12]. The approach has been highly successful and has produced state-of-the-art image analysis models with impressive accuracy.

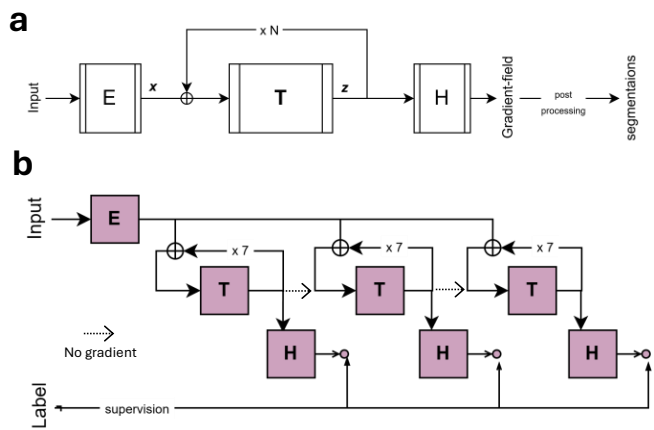

**Figure 1: UCell model architecture.** (**a**) Components of ucell: The embedder module (E) converts 2D input image (at the resolution of 256×256) to feature embeddings, *x* (64×64×d). The core module T is a vision transformer called recursively for multiple times (*N = 21*). The head module (H) project the feature representations from the final iteration output to a 2D gradient field. An offline post-processing routine was employed to convert the gradient field final iteration to single-cell segmentations. (**b**) Training scheme: The full computation graph is unrolled and chunked into three sections, the feature outputs from an earlier chunk were used as constant inputs for the next without gradient propagation. The same features were also fed to the head (H) module to allow independent supervision of each chunk.

The success of the scaling approach, however, has probabaly masked the under-exploration of the alternative: improving the quality of small models. There are plenty of reasons to prefer a small model: (1) Resource – a small model requires less memory to run; (2) A small model is much amendable to test-time scaling [13] strategies and model ensembling [14]; and (3) Large model training often depends on "tribal knowdge", because the model is large enough so that a principal approach to hyperparameter tuning is generally not feasible, whereas for small models easier experimentations may lead to better machnistic understandings.

In this paper, we experiment with training a small vision model, called ucell, to solve the single-cell segmentation problem; single-cell segmentation is relavent to many cell biological research areas, and has attracted a significant amount of research effort to establsh strong prior baselines, making it an impactful problem to focus on. We show here that our model of 10-30M parameters can match the performance of state-of-the-art models an order bigger in size. More importantly, ucell is trained from scratch using only microscopy data [15-18] and does not employ pretraining on a natural image dataset, making the full model building pipeline a significantly simpler one and independent of proprietary resources. UCell achieves this by introducing a recursive structure in its forward computation graph (Fig. 1). More concretely, during forward computation the model makes repeated call of its core transformer module (T, Fig. 1) by feeding the previous output back as the input to the module. This increases the model nonlinearity and feature complexity without significantly ramp up the capacity and number of parameters.

It is worth noting that recursive computing structure is not new in deep learning. Probably the most well-known recursive model is the diffusion model [19], currently the industrial standard for image/video generation and widely adopted in many scientific modeling problems [20]. A diffusion model generates its output by iteratively calling the same model to perform a denoise task, feeding the previous model output as input for the next iteration. Deep equilibrium model is another highly influential modeling architecture [21], which can be viewed as a type of recursion with infinite number of iterations. More recently, recursive structure has also appeared in several reasoning models aimed at solving complex logical puzzles, which exhibited surprising generalizability and parameter efficency [22, 23]. Among them, the TRM [23] model specifically, had a direct influence on the UCell's design, the main difference being that TRM (and other related reasoning models) compute two separate latents as an alternating sequence, whereas ucell has a simpler design and computes only one. We offer some additional rationales in the discussion section of this paper regarding the design choice.

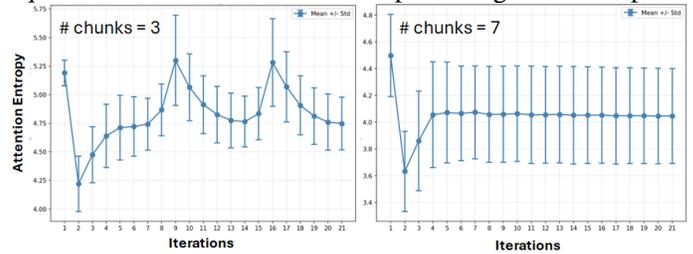

**Figure 2: Regulation of attention entropy.** Number of chunks during training directly affects the attention entropies of the final trained model, and by extension the model quality. We found the optimal number of chunks (3) produced wider exploration of entropy range (left panel). In contrast, dividing the computation graph into too many chunks tends to clamp the attention entropy to a lower value (right panel).

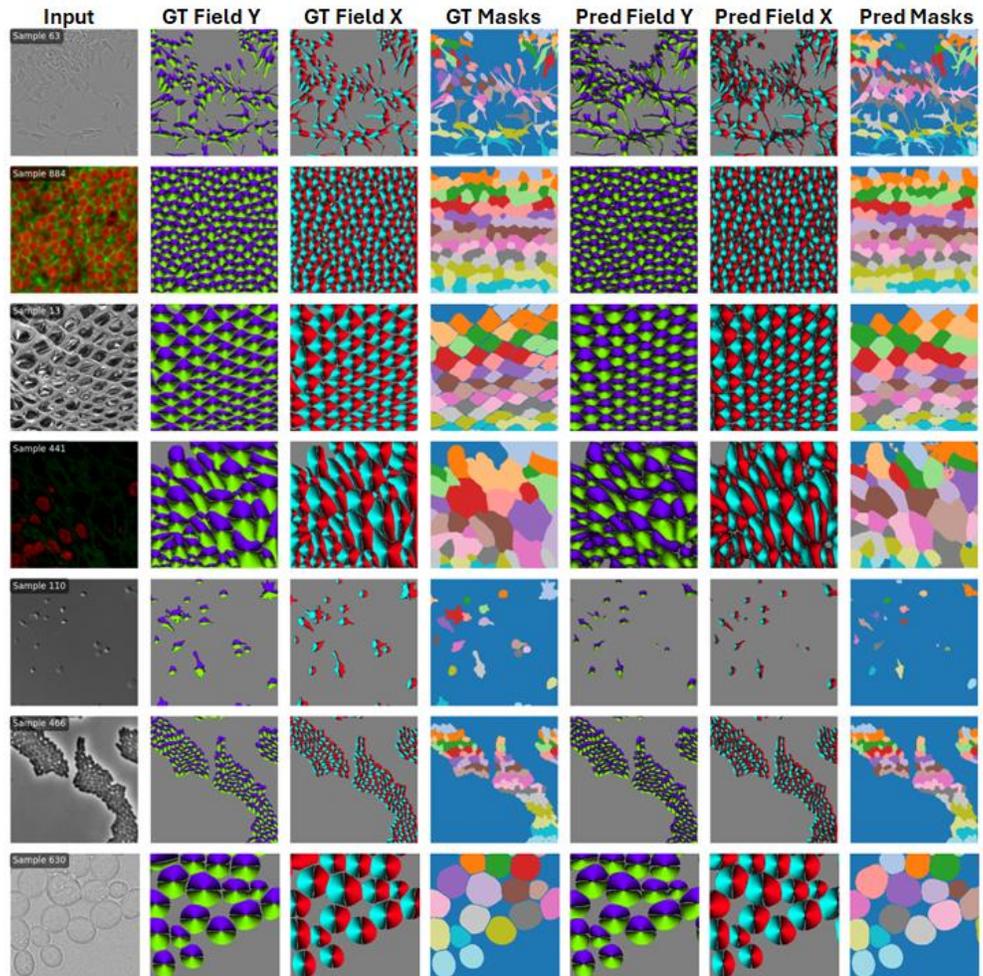

**Figure 3: UCell is a multi-modality single-cell segmentations model.** Test images were randomly selected from the test split of the dataset. The first column shows the input images. The next three representing ground truth based on human labels. The last three columns are ucell predictions. All model predictions were generated using ucell-768.

### Results
**Model architecture and training**
UCell is a small model with three simple components (Fig. 1). The embedder module (E) is a single strided convolution layer tasked to convert the input image to patch-based embeddings. This is a standard approach in almost all modern vision transformers. The head module (H) is a linear probe that does the opposite: projecting the feature representations from the embeddings space to the output space. We use the gradient-field approach, initially proposed in the seminal cellpose paper [24], to represent the output space. This scheme has been well proven to be a very efficient and error-tolerant representation of the segmentations. The core component of ucell is a shallow (two-layers) vision transformer module (T), tasked to process the feature representations with a standard residual connection. The key difference between ucell and a run-of-the-mill vision transformer is that in ucell, instead of feeding forward the output from the module T to H, we sent the data back to be mixed with the input embeddings (additive mixing is used for all our experiments) and to be reprocessed by the T module again, until a certain number of recursion threshold has been reached (21 for our final trained models). A second difference is that we added a small (64) side-band of tokens in the transformer computation to allow carrying additional latent information during forward iterations.

Training of recursive models is far from a settled science; we performed various experiments exploring best strategies for training UCell. Unrolling the full recursion sequence is straightforward but impractical, because it requires a large amount of computer memory in order to store the Jacobian coefficients during gradient computation. A better strategy, we found, is to divide the full sequence into equal size chunks and connect shortcut linkage at the end of each chunk to the head module, thus allowing the training program to supervise each chunk separately. Furthermore, each chunk directly reuses the feature output from the previous chunk as a constant input (i.e., no gradient tracking) allowing memory-efficient gradient computation (Fig. 1).

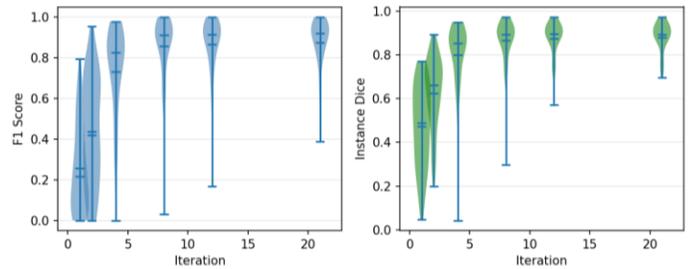

**Figure 5: Iterative refinement of feature representations.** Feature outputs from the T module of ucell were incepted at various specific iterations and were evaluated by projecting to output space using the head (H) module and compared with the ground truth label. Both the F1 scores (left panel) and Dice scores (right panel) demonstrated continued improvements of feature quality. Tests results were from ucell-768 variant and evaluated on test split of cellpose dataset.

The optimal number of chunks were determined empirically by sweeping the hyperparameter (Fig. S1). Surprisingly, the experiments strongly suggested that the chunking method may play an additional regulatory role in model training. In Fig. 2 we compared the attention entropies [25, 26] of a model trained with an optimal number of chunks (m=3) and a model with a higher number of chunks (m=7). The general thinking regarding attention entropy is that it should be banded within a certain range to provide the best results: too low entropy indicates ineffective attention where each token only sees itself, whereas too high entropy represents wash-out of attention and is indicative of a model's inability to focus on the correct context. Evidently, in our experiments, the better model exhibited a broader range of exploration of attention entropies at different iterations, and the supervision point seems to serve as an anchor point pulling down the attention entropy, preventing a wash-out. Conversely, when we increased the number of chunks, the attention entropies were clamped to a low value throughout the computation, resulting in a lesser-quality model.

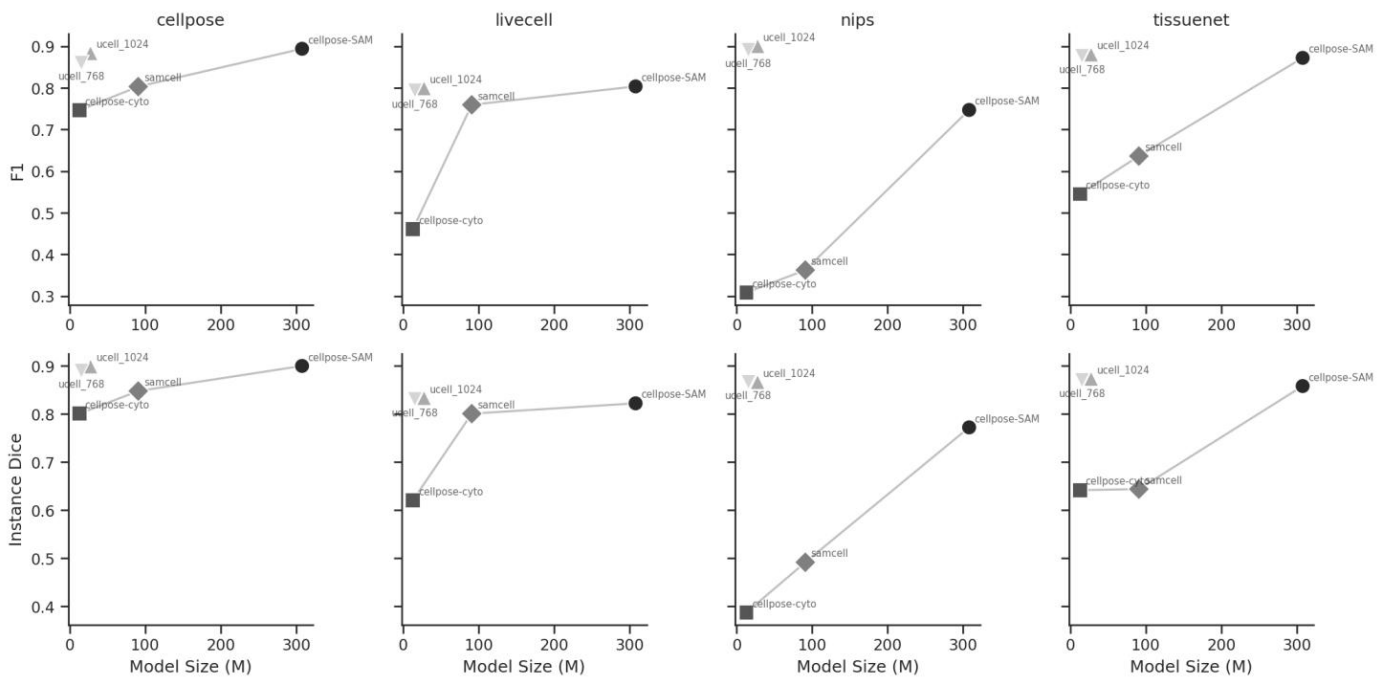

**Figure 4: UCell models are accurate despite the small sizes.** The *F1* scores (top row) and the *Dice* scores were computed using in-domain test datasets. In total, five models were tested, including two ucell variants and three baseline models of different model sizes, measured by the number of model parameters.

We trained two variants of UCell, differing only by the dimensions of the embedding space, called UCell-768 and UCell-1024. The former has ~15 million trainable parameters and the latter is about twice as big (~26 million). The training data encompasses four high-quality manually labeled microscopy datasets (Table S1). For each dataset, ucell also learns a different side-band token initiation, which we found to be sufficient to account for the minor differences in labeling preferences (e.g., incomplete labeling in LiveCell dataset, etc.) without resorting to manual pruning of the training dataset. Unlike most state-of-the-art pipelines, we do not utilize any non-microscopy datasets for pretraining and instead train from scratch using only microscopy data. We employed standard data augmentation techniques, such as random rescaling and random flipping and both model variants were trained at the resolution of 256 × 256, with no additional fine-tuning to higher resolution. See Table S2 for a detailed list of training hyperparameters.

**Model evaluation**

In Fig. 3 we show a set of random examples of ucell-1024 output on a random set of test images. Briefly, the results qualitatively showed that ucell could process a wide variety of input characteristics, including both fluorescence and bright field imaging modality, both mammalian and non-mammalian sample types, to generate high quality instance segmentations results.

To further quantify the ucell's performance, we relied on two metrics: the *F1* score and the instance dice score. The *F1* score is commonly used as a balanced measure of model precision and model recall, defined as:

$$F1 = \frac{2 \times TP}{2 \times TP + FP + FN}$$

where *TP*, *FP* and *FN* denote the number of true positives, false positives, and false negatives, respectively. We consider a detection with at least 0.5 *IOU* (intersection over union) with a ground truth segmentation a *positive* detection, following the field convention. A shortcoming of the *F1* score is that it does not distinguish a low-quality positive detection from a high-quality positive detection, so long as they have an *IOU* value above the 0.5 threshold. To remedy this, we also computed the instance-averaged dice score:

$$D = \frac{1}{2}\left(\frac{\sum_i s_i^p d_i^p}{\sum_i s_i^p} + \frac{\sum_j s_j^g d_j^g}{\sum_j s_j^g}\right)$$

where $s_i^p$ is the area of the *i*-th predicted cell segmentation and $d_i^p$ is the pixel-level *F1* of the *i*-th segmentation against a matched ground-truth segmentation, also known as the segmentation dice. Here the matched ground-truth segmentation is the one that resulted in the highest segmentation dice. The values of $s_j^g$ and $d_j^g$ are similarly defined except that they are for the ground truth segmentations matched against the predictions.

Table 1: model evaluations: in-domain test data

| dataset | model | Size (M) | Accuracy | Recall | F1 | Dice |
|---|---|---|---|---|---|---|
| CellPose | cellpose-cyto | 13 | 82.55 | 68.1 | 74.64 | 80.08 |
| | ucell_768 | 15.7 | 85.45 | 86.59 | 86.01 | 88.99 |
| | ucell_1024 | 27.8 | 88.88 | <u>87.83</u> | 88.36 | 89.9 |
| | samcell | 91 | 84.7 | 76.35 | 80.31 | 84.78 |
| | cellpose-SAM | 308 | <u>92.36</u> | 72.85 | <u>89.39</u> | <u>90</u> |
| LIVECell | cellpose-cyto | 13 | 78.1 | 32.71 | 46.1 | 62.07 |
| | ucell_768 | 15.7 | 77.89 | 80.91 | 79.37 | 83.09 |
| | ucell_1024 | 27.8 | 78.29 | <u>81.62</u> | 79.92 | <u>83.34</u> |
| | samcell | 91 | 80.24 | 72.08 | 75.94 | 80.08 |
| | cellpose-SAM | 308 | <u>88.65</u> | 72.85 | <u>80.36</u> | 82.23 |
| NIPS | cellpose-cyto | 13 | 27.13 | 35.74 | 30.85 | 38.75 |
| | ucell_768 | 15.7 | 89.28 | 88.95 | 89.12 | 86.65 |
| | ucell_1024 | 27.8 | <u>90.91</u> | <u>89.21</u> | <u>90.05</u> | <u>86.66</u> |
| | samcell | 91 | 29.08 | 48.14 | 36.26 | 49.18 |
| | cellpose-SAM | 308 | 65.18 | 85.63 | 74.71 | 77.23 |
| TissueNet | cellpose-cyto | 13 | 78.3 | 41.81 | 54.51 | 64.18 |
| | ucell_768 | 15.7 | 89.83 | 85.6 | 87.66 | 87.06 |
| | ucell_1024 | 27.8 | 90.02 | <u>86.12</u> | <u>88.03</u> | <u>87.33</u> |
| | samcell | 91 | 74.29 | 41.96 | 63.63 | 64.39 |
| | cellpose-SAM | 308 | <u>91.91</u> | 82.76 | 87.21 | 85.82 |

Table 2: model evaluations: out-of-domain test data

| dataset | model | Size (M) | Accuracy | Recall | F1 | Dice |
|---|---|---|---|---|---|---|
| MUSC | cellpose-cyto | 13 | <u>20.11</u> | 43.11 | <u>27.42</u> | 26.71 |
| | ucell_768 | 14.7 | 6.402 | 61.17 | 11.59 | 37.72 |
| | ucell_1024 | 26.4 | 7.054 | 58.45 | 12.59 | 36.61 |
| | samcell | 91 | 12.48 | <u>76.89</u> | 21.48 | 38.35 |
| | cellpose-SAM | 308 | 16.92 | 71.46 | 27.36 | <u>43.59</u> |
| Ovules | cellpose-cyto | 13 | 85.08 | 45.05 | 58.91 | 73.41 |
| | ucell_768 | 14.7 | 81.37 | 57.47 | 67.36 | 81.97 |
| | ucell_1024 | 26.4 | 79.9 | <u>59.4</u> | 68.14 | <u>83.04</u> |
| | samcell | 91 | 81.25 | 50.99 | 62.66 | 76.93 |
| | cellpose-SAM | 308 | <u>90.86</u> | 55.11 | <u>70.77</u> | 82.63 |
| PBL_HEK | cellpose-cyto | 13 | 77.37 | 17.2 | 36.48 | 46.13 |
| | ucell_768 | 14.7 | 51.86 | <u>71.78</u> | <u>66.26</u> | 64.03 |
| | ucell_1024 | 26.4 | 66.9 | 44.82 | 53.67 | 69.96 |
| | samcell | 91 | 71.52 | 51.69 | 60.8 | <u>71.98</u> |
| | cellpose-SAM | 308 | <u>86.96</u> | 46.56 | 63.63 | 70.44 |
| PBL_N2A | cellpose-cyto | 13 | 91.75 | 80.46 | 85.92 | 78.65 |
| | ucell_768 | 14.7 | 88.7 | 85.41 | 81.82 | <u>91.5</u> |
| | ucell_1024 | 26.4 | 92.09 | 86.29 | 89.1 | 82.69 |
| | samcell | 91 | 91.37 | <u>88.97</u> | <u>90.16</u> | 85.67 |
| | cellpose-SAM | 308 | <u>95.48</u> | 83.84 | 89.47 | 82.77 |

We performed both in-domain tests (Fig 4, Table 1), using the test or validation splits from the same datasets used for training (Table S1), as well as out-of-domain tests (Table 2, Fig. S2), using orthogonal datasets prepared independently of the training datasets (Table S1). Three high-quality segmentation models, cellpose-sam [8], samcell (generalist variant) [12] and cellpose (cyto3 variant) [24], served as our baselines. The first two are foundation models created using the pre-train + fine-tune pipelines, and the last one is a very well-known small model with a significant level of deployment in research labs. We choose not to use any per-image metadata, such as cell sizes, in the test. The choice is slightly unfair to the cellpose model, which is designed to leverage such information to boost model accuracy, but allows a more consistent comparison among different models and different datasets.

For the in-domain tests, we found the ucell variants perform significantly better than both cellpose and samcell, and was competitive against cellpose-sam, the best of our baselines. Specifically, UCell-1024 exhibited slightly lower accuracy on the cellpose dataset in comparison to cellpose-sam but performed slightly better on the TissueNet dataset. On LiveCell tests, cellpose-sam is better at detection (higher *F1*) but ucell won on quality (higher *dice*). UCell is also significantly stronger than cellpose-sam on NIPS challenge dataset, but this comparison is slightly unfair because cellpose-sam had not been trained on the full training set of NIPS data due to their manual data pruning.

As shown in Fig. 4, excluding ucell, the rest of the models exhibited the expected positive correlation between model size and the model accuracy, with cellpose-sam being the best model as well as the largest model. In contrast, the evaluation results from ucell sit in a location deviated from this trend line. To have a better understanding of ucell's computation path, we intercepted its intermediate feature outputs at selected iteration and evaluated the accuracy of the intermediate results (Fig. 5). A general take-home from this exercise is that UCell continuously improve the output quality through every iteration of T module processing. The average metrics increased during the first two thirds of the iterations, whereas the last one third of the iterations seemed to be dedicated to dealing with outlier and edge cases, which ensured the consistency on all types of inputs.

Test results on out-of-domain images were much noisier (Table 2, Fig. S2), with no single model being the clear winner. Every model listed here, including cellpose, the smallest among these, was found to be ranked at the top in at least one category. The lesson here, in our opinion, is that despite the expensive pre-training pipeline, today's vision foundation models have not yet solved the generalization problem in a satisfactory manner. Instead, the models remain somewhat brittle and can break catastrophically in an unexpected manner. Case in point, all five models evaluated here performed quite poorly in the MUSC dataset (part of the cell tracking consortium collection [27]). The best performer could only achieve F1 scores in the twenties, far from being useful in real data analysis. On that note, we recognize that the ucell also has limited generalizability. In fact, it performed at the bottom of the list on MUSC images, although it scored better in the three other tests.

**Few-shot model adaptation.**
The limited generalizability exiting segmentation models motived us to further examine UCell's adaptability to new data domain. Our hypothesis is that its basal quality combined with its small size would facilitate model adaptation. Here we will focusing on the one-shot and few-shot model adaptations, which represent more practical workflows for biomedical researchers. We explored the following adaptation scenarios as case studies (Fig. 6):

*TissueNet Nuclei*: The TissueNet dataset [15] comes with two sets of ground truth labels: one for the cells and one for the nuclei. The ucell is trained with the former but here we tested whether it can switch to nuclei segmentation with a few training examples. The experiment provides insights about the more general case of adapting new segmentation preferences.

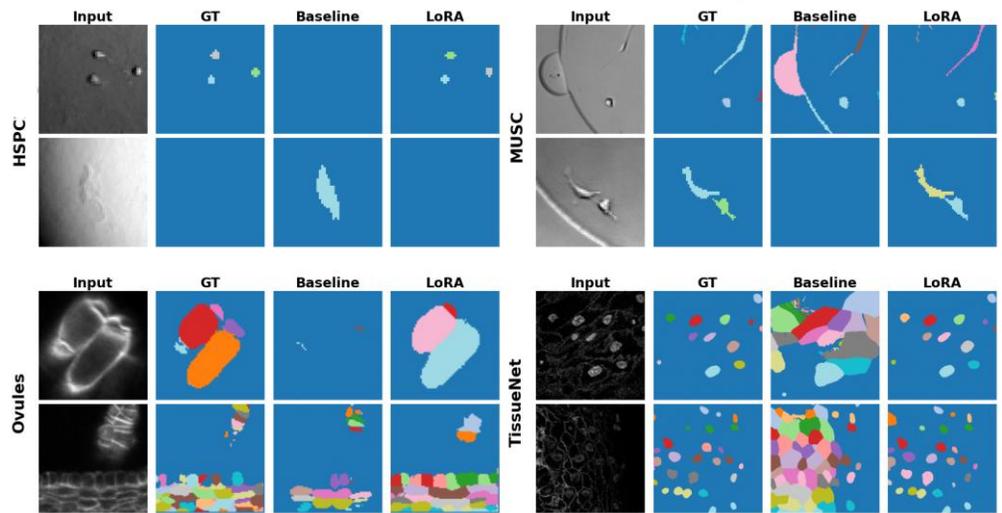

**Figure 6: UCell is readily adaptable to new data characteristics using few-shot finetuning.** Representative examples of ucell-768 segmentation on various test images either before (baseline) or after few-shot adaptation using the LoRA finetuning pipeline (LoRA). Both the input image (input) and ground truth segmentation (GT) are also shown for references.

*MUSC*: One of our out-of-domain test set [27]. UCell appears to struggle with this dataset due to unseen image features: novel cellular features resulted in false negatives and confusing background features resulted in false positives.

*HSPC*: A mixture of the two previous cases. The HSPC dataset [28] is comprised of images of primary human differentiating hematopoietic progenitor cells differentiating and committing to myeloid lineages. The small cell sizes combined with low cell density confused the model mistaking them as debris. In addition, the labeled cell boundaries were very imprecise because the original work's focus was lineage tracking not segmentation. A successful adaptation needs to resolve both issues.

*Ovules*: A semi-automatically labeled 3D segmentation dataset [17] exhibiting novel image features relative to Ucell's training data, which are almost entirely based 2D imaging modalities. For evaluation, we treat the data as stacks of 2D segmentations, ignoring the 3D nature of the input data, thus allowing comparison with the rest of the benchmarks in this study.

We performed all experiments using the trained UCell-768 as the base model. For each trial, one (one-shot) or four (few-shot) image/label pairs were randomly chosen from the dataset to be training examples, except for the ovules dataset, for which we used the center slice of the stack. To reduce sampling noise, images with cell counts below average were removed from the selection pool. We then fine-tuned the model using the new training examples. Importantly, we employed no early stopping criteria or any other measures to prevent overfitting. We consider this a notable feature of our training pipelines, because in practice, monitoring overfitting can be difficult without a full validation dataset. Instead, in our experiments, we simply train the model until training loss convergence. EMA is applied throughout the training course to obtain a more stable checkpoint, which is then used directly for evaluation.

Two pipelines were tested: (1) We train the full model with no frozen parameters. (2) We train the model via a LoRA (low rank adaptation) scheme [29], which reduces the number of trainable parameters to ~0.27M at LoRA rank 16.

Fig. 7 summarizes the quantitative results. The data confirms that ucell model can be readily adapted to new data characteristics. For the MUSC dataset, for example, LoRA based few shot adaptation increased the *F1* score from 0.12 to 0.88±0.02(s.d.). In majority of the cases, the LoRA fine tuning pipeline performed better than full model fine tuning, although the differences are small, especially in the few-shot cases.

The experiments on the ovules dataset achieved the least amount of improvement in model accuracy. We tentatively attributed this to two factors: (1) The baseline model already performed relatively well on this dataset, leaving less room for improvements. (2) The semi-automatically generated segmentation labels have many errors, the characteristics of which might be difficult to generalize from a few training images. This sets an upper bound on how well a model can perform on this dataset.

## Discussion
### A connection to expectation maximization (EM)

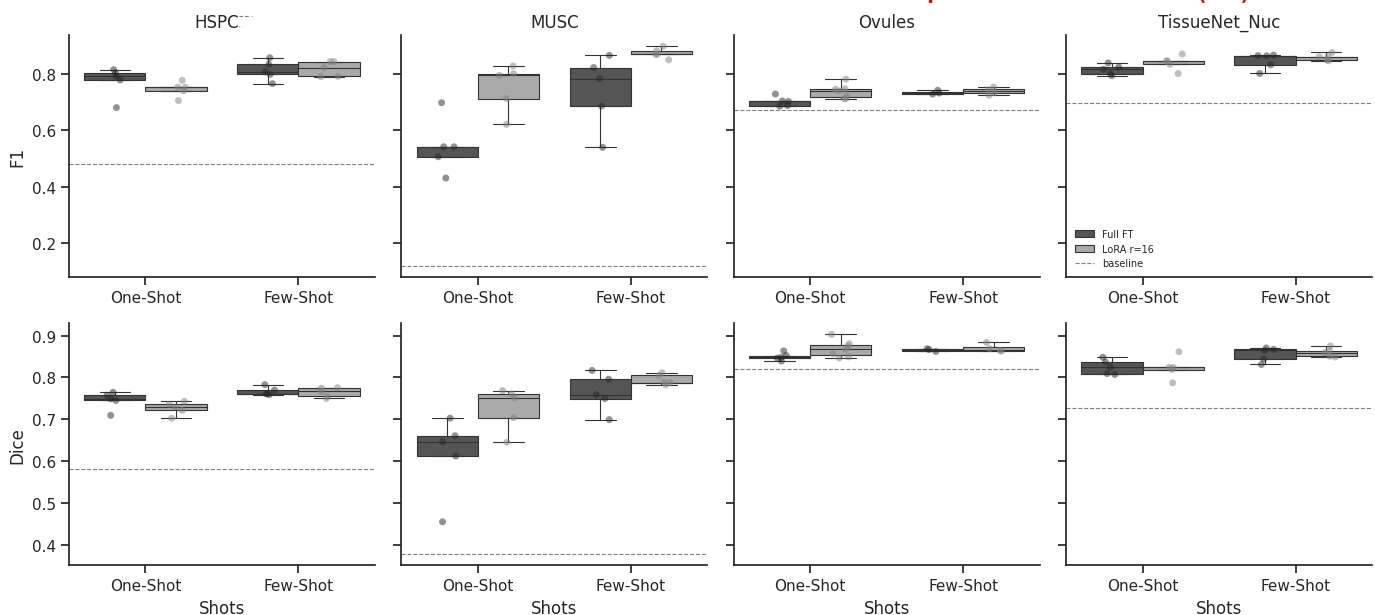

**Figure 7: Quantitative evaluation of the few-shot adaptation pipelines.** F1 score (top) and instance dice scores (bottom) were macro-average of the full dataset. Each trial was repeated for five times to gauge sampling noise and consistency of the pipeline. Dashlines in each panel represents the starting point of the relevant metric before model adaptation. All results were from ucell-768.

It is already stated that ucell's recursive structure resembles those of recent reasoning models, such as HRM/TRM, which were designed to solve logic puzzles, such as Sudoku and 2D mazes, in a parameter-efficient manner. Both HRM and TRM compute two separate latents, $z_L^i$ and $z_H^i$, as an alternating sequence, i.e.,

$$z_L^i = f_L(z_L^{i-1};\ x, z_H^{i-1})$$
$$z_H^i = f_H(z_H^{i-1};\ z_L^i)$$

where $x$ is the input embeddings and the model output is a transformation of $z_H$: $y = h\left(z_H^{i_{final}}\right)$. In contrast, the ucell's recursion structure is simpler:

$$z^i = f_{HL}(z^{i-1};\ x)$$

which can be viewed as compacting the two functional steps $f_H$ an $f_L$ into a single learned function $f_{HL}$.

While the original authors of the HRM/TRM papers did not dive deep into the rationales of recursive model design, it is not difficult to see that the recursion structures have a clear parallel to the classical expectation maximization (EM) method [30], which iteratively solve an inference problem of the structure:

$$X \to Z \to Y$$

where $X$ is the observation, $Z$ is an unobserved latent, and $Y$ is the variable of interest. EM finds the maximum likelihood estimation of $Y$ by performing iterative optimization:

$$Z^i \sim P_Z(\cdot\ |X, Y^{i-1})$$
$$Y^i = \arg\max_y \mathbb{E}_{Z \sim Z^i}[\mathcal{L}(Y; Z)]$$

where $\mathcal{L}(\cdot)$ is the log likelihood function. The power of the EM algorithm stem from the fact that in each iteration, it performs computation of the full tentative distribution of the latent $Z$, and the next estimation of $Y$ is based on the knowledge of this full distribution. It is intuitive to see why this is important for solving puzzles like 2D maze, which requires the tracking of many intermediate possibilities during the process. Indeed, in HRM, the main part of the compute is devoted to the $f_L(\cdot)$ function, which is analogous to obtaining conditional distribution of $P_Z(\cdot\ |X, Y^{i-1})$ in the EM algorithm. Their model design therefore force the model to put more focus on learning this distribution function during training.

Even though the cell segmentation problem can also be formulated as an EM problem, it seems reasonable to argue that its latent space is significantly simpler than that of Sudoku and mazes, i.e., a human doing segmentation would typically require less backtracking to finish the task. This seems to be an important reason that our model, using a much simpler iterative structure, still solves the problem well.

### A case for the small models

While there is no doubt that scaling is a powerful tool for building "smarter" machine learning models, and will remain a key focus of future research, we feel the sole emphasis on scaling may not be healthy in the long term and want to also provide an argument for researching small models, especially for biology and biomedical problems. First of all, the environmental externalities [31, 32] of deep learning research have been extensively documented and discussed, which we won't repeat. In addition, model scaling requires more training data, which is expensive to compile. For the vision models, researchers have recently turned to commercial entities for such datasets, and in some cases, researcher simply reused pretrained large foundation models. So far, the industry has been very willing to share their data and model weights in the public domain. But it should be pointed out that there is no guarantee that this will remain so in the future, i.e., cue the trajectory of language model development history. On top of that, the copyright issue of datasets used in commercial models remains murky [33, 34].

Part of the reason that the field has put so much emphasis on scaling may stem from the dogmatic view that scaling is the only possible path to an increased model generalizability and high quality models. The small amount of data provided in this paper is not enough to disprove the dogma, but hopefully serves to attract more research interests in similar problems and approaches. After all, it is worth remembering that despite tremendous advancements and improvement in modern neural-nets, the generalizability gap between machine and human remains – humans are still much more capable of generalizing from a small numer of examples during learning. In that sense, the modern training pipeline of relying on massive data corpus is a crutch to workaround the generalizbility gap, which is a fundamental problem yet to be solved.

### References

1. Kaplan, J. et al. Scaling Laws for Neural Language Models. arXiv:2001.08361 (2020).
2. Hoffmann, J., Borgeaud, S., Mensch, A. et al. Training Compute-Optimal Large Language Models. Advances in Neural Information Processing Systems 35 (NeurIPS 2022).
3. Wei, J., Tay, Y., Bommasani, R. et al. Emergent Abilities of Large Language Models. arXiv:2206.07682 (2022).
4. Khan, A., Viana, M. P., Mishra, S. & Manjunath, B. S. Bright 4B: Scaling Hyperspherical Learning for Segmentation in 3D Brightfield Microscopy. arXiv:2512.22423 (2025).
5. Zeng, Y., Xie, J., Shangguan, N. et al. scLong: A Billion-Parameter Foundation Model for Capturing Long-Range Gene Context in Single-Cell Transcriptomics. Nature Communications (2026).
6. Kirillov, A., Mintun, E., Ravi, N. et al. Segment Anything. Proceedings of the IEEE/CVF International Conference on Computer Vision (ICCV) 2023. arXiv:2304.02643.
7. Ravi, N., Gabeur, V., Hu, Y.-T. et al. SAM 2: Segment Anything in Images and Videos. arXiv:2408.00714 (2024).
8. Stringer, C., Wang, T. & Pachitariu, M. Cellpose-SAM: superhuman generalization for cellular segmentation. bioRxiv 2025.04.28.651001 (2025).
9. Shao, A., Mohanty, R., Wang, Y. et al. CellSAM: a foundation model for cell segmentation. Nature Methods (2025).

10. Zhu, J., Liu, J., Zhang, Z. et al. Medical SAM 2: Segment medical images as video via Segment Anything Model 2. arXiv:2408.00874 (2024).
11. Zhu, L., Chen, T., Yang, D. et al. SAM2-Adapter: Evaluating & Adapting Segment Anything 2 in Downstream Tasks. arXiv:2408.04579 (2024).
12. VandeLoo, A. D., Malta, N. J., Sanganeriya, S. et al. SAMCell: Generalized label-free biological cell segmentation with segment anything. PLOS One (2025).
13. Wang, P., Li, L., Shao, Z. et al. s1: Simple Test-Time Scaling. arXiv:2501.19393 (2025).
14. Zhang, Y., Yang, J., Yuan, Y. et al. Ensemble Deep Learning: A Review. arXiv:2104.02395 (2021).
15. Naylor, J., Li, K., Schürch, C. et al. Whole-cell segmentation of tissue images with human-level performance using large-scale data annotation and deep learning. Nature Biotechnology 40, 555–564 (2022).
16. Edlund, C., Jackson, T. R., Khalid, N. et al. LIVECell—A large-scale dataset for label-free live cell segmentation. Nature Methods 18, 1038–1045 (2021).
17. Wolny, A., Cerrone, L., Vijayan, A. et al. Accurate and versatile 3D segmentation of plant tissues at cellular resolution. eLife 9, e57613 (2020).
18. Ma, J., Xie, R., Ayyadhury, S. et al. The Multi-modality Cell Segmentation Challenge: Towards Universal Solutions. Nature Methods (2024).
19. Ho, J., Jain, A. & Abbeel, P. Denoising Diffusion Probabilistic Models. arXiv:2006.11239 (2020).
20. Song, Y., Shen, L., Wang, L. et al. Diffusion Models in Medical Imaging: A Comprehensive Survey. Medical Image Analysis 88, 102868 (2023).
21. Bai, S., Kolter, J. Z. & Koltun, V. Deep Equilibrium Models. Advances in Neural Information Processing Systems 32 (NeurIPS 2019).
22. Wang, G., Li, J., Sun, Y. et al. Hierarchical Reasoning Model. arXiv:2506.21734 (2025).
23. Jolicoeur-Martineau, A., Li, K., Piché-Taillefer, R. et al. Less is More: Recursive Reasoning with Tiny Networks. arXiv:2510.04871 (2025).
24. Stringer, C., Wang, T., Michaelos, M. & Pachitariu, M. Cellpose: a generalist algorithm for cellular segmentation. Nature Methods 18, 100–106 (2021).
25. Zhai, S., Wang, H., Liu, H. et al. Stabilizing Transformer Training by Preventing Attention Entropy Collapse. Proceedings of Machine Learning Research 202, 40221-40241 (2023).
26. Liu, L., Liu, X., Gao, J. et al. Understanding the Difficulty of Training Transformers. EMNLP 2020.
27. Maška, M., Ulman, V., Potěšil, D. et al. The Cell Tracking Challenge: 10 years of objective benchmarking. Nature Methods 20, 1010–1020 (2023).
28. Garcia, C., de Laval, B., Godin, C. et al. Multiparameter analysis of timelapse imaging reveals kinetics of megakaryocytic erythroid progenitor clonal expansion and differentiation. Scientific Reports 12, 15293 (2022).
29. Hu, E. J., Shen, Y., Wallis, P. et al. LoRA: Low-Rank Adaptation of Large Language Models. arXiv:2106.09685 (2021).
30. Dempster, A. P., Laird, N. M. & Rubin, D. B. Maximum Likelihood from Incomplete Data via the EM Algorithm. Journal of the Royal Statistical Society: Series B (Methodological) 39, 1–22 (1977).
31. Strubell, E., Ganesh, A. & McCallum, A. Energy and Policy Considerations for Deep Learning in NLP. Proceedings of the 57th Annual Meeting of the Association for Computational Linguistics (ACL) 3645-3650 (2019).
32. Luccioni, A. S., Viguier, S. & Ligozat, A.-L. Estimating the Carbon Footprint of BLOOM, a 176B Parameter Language Model. Journal of Machine Learning Research 24, 1-15 (2023).
33. U.S. Copyright Office. Copyright and Artificial Intelligence: Part 3 — Generative AI Training. (2025). https://www.copyright.gov/ai/
34. Thomson Reuters Co. v. ROSS Intelligence Inc. (D. Del. 2025).

## Acknowledgements
We thank J. Li for critical reading of the manuscript and for helpful discussions.

## Competing interest statement
We claim no competing interest.

**Table S1 Dataset used in the study**

**Training**

| | Dataset | Modality | # images | # cells | Ref |
|---|---|---|---|---|---|
| | Cellpose | Mixed | 796 | 110,130 | 1 |
| | LiveCell | Bright field | 3253 | 1,018,576 | 2 |
| | TissueNet | Fluorescence | 2580 | 988,150 | 3 |
| | NIPS challenge | Mixed | 1000 | 168,364 | 4 |

**Test / Adaptation**

| | Dataset | Modality | # images | # cells | Ref |
|---|---|---|---|---|---|
| In-domain | Cellpose | Mixed | 68 | 7,201 | 1 |
| | LiveCell | Bright field | 1564 | 462,261 | 2 |
| | TissueNet | Fluorescence | 1324 | 145,222 | 3 |
| | NIPS challenge | Mixed | 100 | 45,445 | 4 |
| out-of-domain | Ovules (test) * | Fluorescence | 408 | 70,392 | 5 |
| | PBL-HEK | Bright field | 5 | 1,948 | 6 |
| | PBL-N2A | Bright field | 5 | 1,714 | 6 |
| | MUSC† | Bright field | 100 | 515 | 7 |
| | TissueNet (Nuclei) | Fluorescence | 1324 | 135,633 | 3 |
| | HSPC | Bright field | 421 | 2,446 | 8 |

* reduced along z axis by 5x to avoid highly similar inputs

† adpated from the gold-standard segmentations of the BF-C2DL-MUSC set in CTC (cell tracking consorsium)

**Table S2 Training hyperparameters**

| Parameter | Value |
|---|---|
| *General* | |
| Training Resolution | 256 x 256 |
| Embedding stride | 4 |
| Max recursion | 21 |
| Num chunks | 3 |
| Batch size | 196 |
| Num epochs | 500 |
| EMA | 0.999 |
| | |
| *Augmentation* | |
| Resize (S) | $Log(S) \sim N(0, 0.6)$ |
| Aspect ratio (A) | $Log(A) \sim N(0, 0.2)$ |
| Flip | Horizontal + Vertical |
| | |
| *Optimization* | |
| Optimizer | adam [9] |
| Learning Rate | $0.001 \rightarrow 0.0001$ |
| Weight decay [10] | 1.0 |
| beta_1 | 0.9 |
| beta_2 | 0.95 |

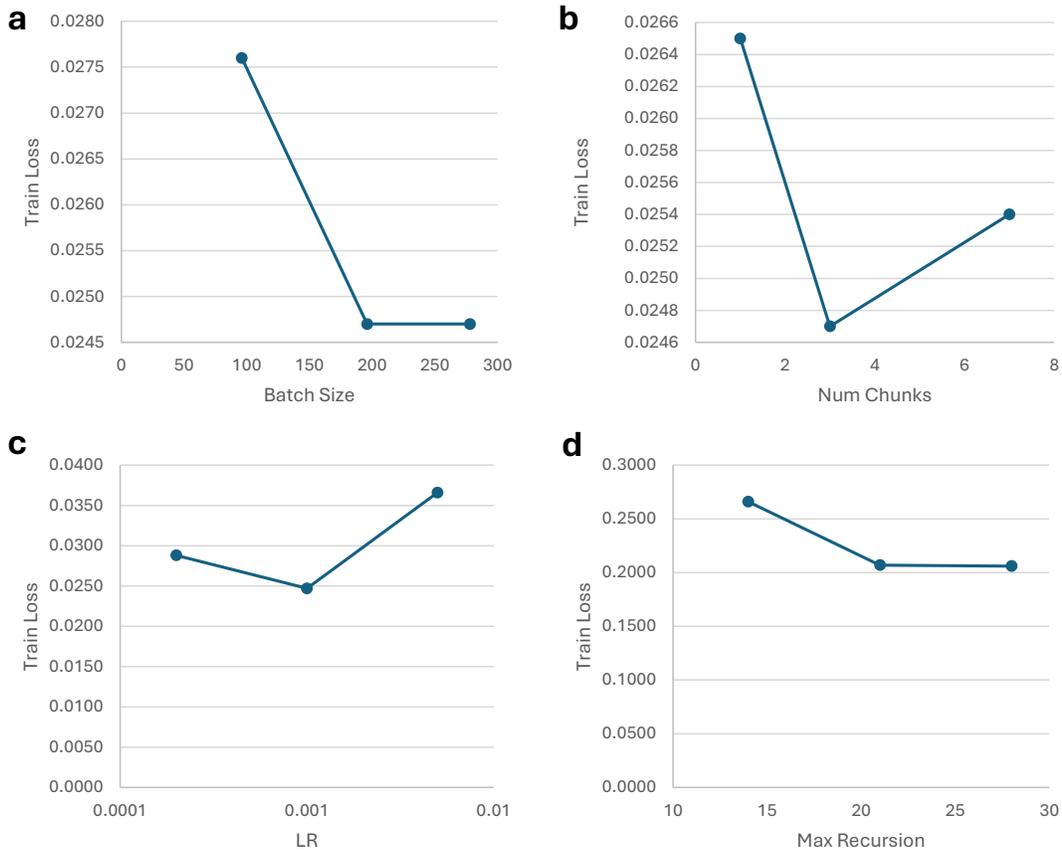

**Figure S1 Hyperparameter optimizations.** We performed limited hyperparameter sweep to investigate their impact on model training. Main results are shown above. Training losses instead of testing metrics were used as a surrogate for training efficiency, as they are more stable. Loss values were sampled at 250[th] epoch except for (d), which was from the 400[th] epoch.

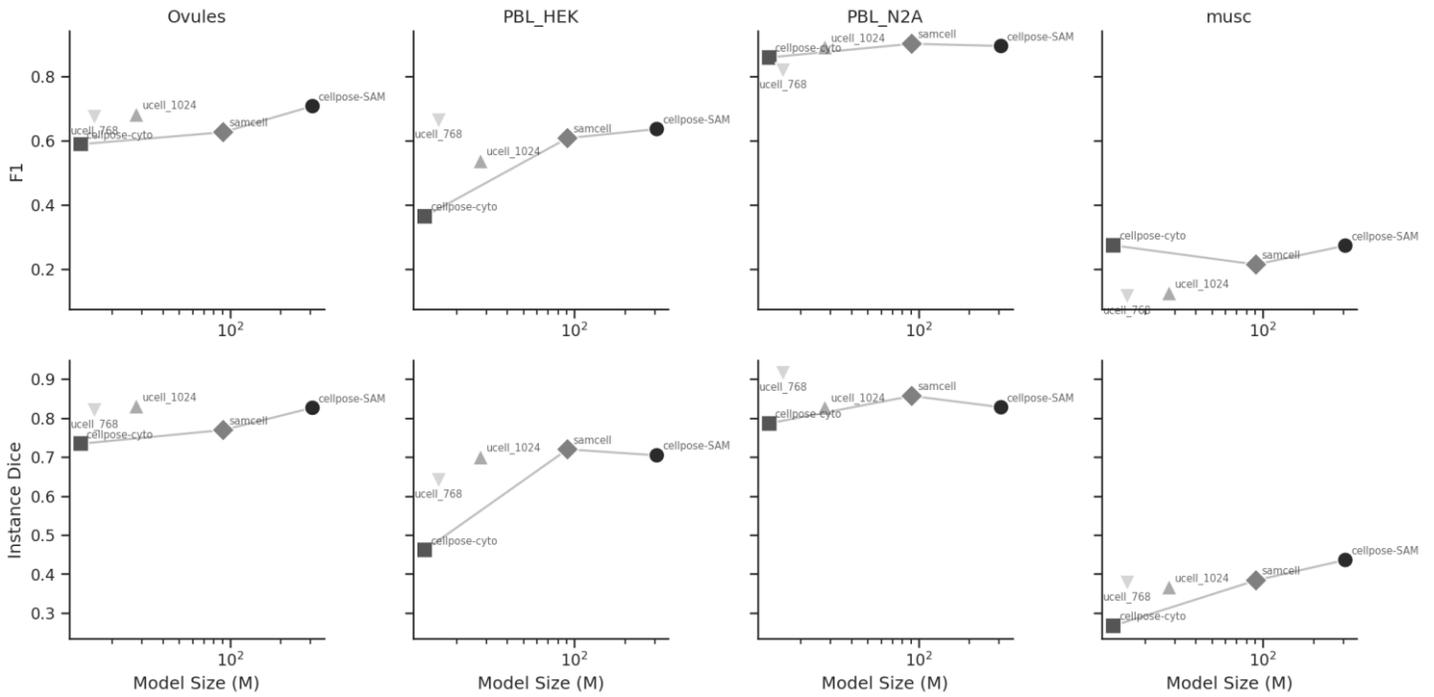

**Figure S2 Out-of-domain model evaluations.** Evaluation on novel data characteristics demonstrated limited model generalizability for all models tested in this study.

Supplementary References:

S1. Stringer, C., Wang, T., Michaelos, M. & Pachitariu, M. Cellpose: a generalist algorithm for cellular segmentation. Nature Methods 18, 100–106 (2021).
S2. Edlund, C., Jackson, T. R., Khalid, N. et al. LIVECell—A large-scale dataset for label-free live cell segmentation. Nature Methods 18, 1038–1045 (2021).
S3. Naylor, J., Li, K., Schürch, C. et al. Whole-cell segmentation of tissue images with human-level performance using large-scale data annotation and deep learning. Nature Biotechnology 40, 555–564 (2022).
S4. Ma, J., Xie, R., Ayyadhury, S. et al. The Multi-modality Cell Segmentation Challenge: Towards Universal Solutions. Nature Methods (2024).
S5. Wolny, A., Cerrone, L., Vijayan, A. et al. Accurate and versatile 3D segmentation of plant tissues at cellular resolution. eLife 9, e57613 (2020).
S6. VandeLoo, A. D., Malta, N. J., Sanganeriya, S. et al. SAMCell: Generalized label-free biological cell segmentation with segment anything. PLOS One (2025).
S7. Maška, M., Ulman, V., Potěšil, D. et al. The Cell Tracking Challenge: 10 years of objective benchmarking. Nature Methods 20, 1010–1020 (2023).
S8. Garcia, C., de Laval, B., Godin, C. et al. Multiparameter analysis of timelapse imaging reveals kinetics of megakaryocytic erythroid progenitor clonal expansion and differentiation. Scientific Reports 12, 15293 (2022).
S9. Kingma, D. P. & Ba, J. Adam: A Method for Stochastic Optimization. arXiv:1412.6980 (2014).
S10. Loshchilov, I. & Hutter, F. Decoupled Weight Decay Regularization. arXiv:1711.05101 (2017).

†